%% file: 0_main.tex
\documentclass[11pt,letterpaper,logo,twocolumn]{style}

\usepackage[numbers]{natbib}
\usepackage{graphicx}
\usepackage{booktabs}
\usepackage{amsmath,amsfonts,amssymb}
\usepackage{subcaption}
\usepackage{multirow}
\usepackage{colortbl}
\usepackage{listings}
\usepackage{xparse}
\usepackage{fontawesome5}
\usepackage{threeparttable}

\graphicspath{{./}{fig/}{figures/}{plot/}{pdf/}{table/}}
\usepackage{amsthm}
\tcbuselibrary{skins,breakable}
\tcbuselibrary{listingsutf8}
\usepackage{titletoc}
\usepackage{pifont}
\usepackage{mathtools}
\usepackage{bbm}
\usepackage{makecell}
\usepackage{adjustbox}

\usepackage{amsmath}
\usepackage{comment}
\usepackage{booktabs}
\usepackage{tabularx}
\usepackage{ragged2e}
\usepackage{pifont}

\usepackage{times}
\usepackage{latexsym}
\usepackage[edges]{forest}
\usepackage[T1]{fontenc}
\usepackage[utf8]{inputenc}
\usepackage{microtype}
\usepackage{inconsolata}
\usepackage{tcolorbox}
\usepackage{subcaption}
\usepackage{lipsum}
\usepackage{multirow}
\usepackage{array}
\newcolumntype{C}[1]{>{\raggedright\arraybackslash}p{#1}}
\definecolor{softpurple}{RGB}{229,224,236}  
\definecolor{softrose}{RGB}{244,224,232} 
\definecolor{softpink}{RGB}{248,226,234} 
\definecolor{softred}{RGB}{242,220,219}     
\definecolor{softcoral}{RGB}{250,224,222} 
\definecolor{softpeach}{RGB}{255,230,220} 
\definecolor{softorange}{RGB}{255,236,214} 
\definecolor{softgold}{RGB}{249,239,206} 
\definecolor{softyellow}{RGB}{255,242,204}  
\definecolor{softbeige}{RGB}{240,232,220} 
\definecolor{softsand}{RGB}{244,238,230} 
\definecolor{softolive}{RGB}{235,241,222} 
\definecolor{softsage}{RGB}{229,239,224} 
\definecolor{softgreen}{RGB}{226,239,218}   
\definecolor{softmint}{RGB}{221,242,234} 
\definecolor{softteal}{RGB}{208,232,230} 
\definecolor{softcyan}{RGB}{220,245,245} 
\definecolor{softperiwinkle}{RGB}{224,230,246}
\definecolor{softsky}{RGB}{217,235,247} 
\definecolor{softblue}{RGB}{220,230,242}    
\definecolor{softbluegray}{RGB}{226,232,238} 
\definecolor{softgray}{RGB}{240,240,240}     
\definecolor{magenta}{RGB}{227, 0, 140}
\definecolor{orange}{RGB}{234, 67, 0}

\newcommand{\revise}[1]{}

\newtcolorbox{takeaway}[1]{
    colback=softbluegray,    
    colframe=softbluegray,  
    fonttitle=\bfseries,      
    coltitle=black,           
    attach title to upper,    
    after title={\ },         
    sharp corners,            
    boxrule=0.pt,             
    left=5pt, right=5pt, top=5pt, bottom=5pt, 
    title={#1},               
    fontupper=\small
}

\title{A Survey on Agent-as-a-Judge}
\runningtitle{A Survey on Agent-as-a-Judge}
\PublicDate{2026-1-8}

\author{%
  {\Authfont
    \textbf{Runyang You}\equal\textsuperscript{1} \quad
    \textbf{Hongru Cai}\equal\textsuperscript{1} \quad
    \textbf{Caiqi Zhang}\textsuperscript{2} \quad
    \textbf{Qiancheng Xu}\textsuperscript{1} \quad 
    \textbf{Meng Liu}\textsuperscript{3} \quad
    \textbf{Tiezheng Yu}\textsuperscript{4} \quad
    \textbf{Yongqi Li}\advisor\textsuperscript{1} \quad
    \textbf{Wenjie Li}\textsuperscript{1} \quad
  }\\
  {\Affilfont
    \textsuperscript{1} The Hong Kong Polytechnic University \quad
    \textsuperscript{2} University of Cambridge \quad \\
    \textsuperscript{3} Shandong Jianzhu University \quad
    \textsuperscript{4} Huawei Technologies \quad
    \\
    \equal\ Equal contribution \quad \advisor\ Corresponding author \\
    \texttt{runyang.y@outlook.com,
\{henry.hongrucai, liyongqi0\}@gmail.com,
cswjli@comp.polyu.edu.hk}
  }
}

\begin{document}

\begin{abstract}
\textit{LLM-as-a-Judge} has revolutionized AI evaluation by leveraging large language models for scalable assessments. However, as evaluands become increasingly complex, specialized, and multi-step, the reliability of \textit{LLM-as-a-Judge} has become constrained by inherent biases, shallow single-pass reasoning, and the inability to verify assessments against real-world observations. This has catalyzed the transition to \textit{Agent-as-a-Judge}, where agentic judges employ planning, tool-augmented verification, multi-agent collaboration, and persistent memory to enable more robust, verifiable, and nuanced evaluations. Despite the rapid proliferation of agentic evaluation systems, the field lacks a unified framework to navigate this shifting landscape. To bridge this gap, we present the first comprehensive survey tracing this evolution. Specifically, we identify key dimensions that characterize this paradigm shift and establish a developmental taxonomy. We organize core methodologies and survey applications across general and professional domains. 
Furthermore, we analyze frontier challenges and identify promising research directions, ultimately providing a clear roadmap for the next generation of agentic evaluation.
\end{abstract}

\newcommand{\TitleLinks}{%
\centering
    \vspace{6pt}
    {\noindent\absfont\fontsize{11}{13}\selectfont
    \faGithub\ Project Page: \url{https://github.com/ModalityDance/Awesome-Agent-as-a-Judge}\par}%
}


\maketitle

\input{1_intro}

\input{2_background}
\input{3_taxonomy}

\input{4_apps}

\input{5_discussion}

\input{6_conclusion}


\bibliographystyle{unsrtnat} 
\bibliography{ref}


\appendix

\end{document}

%% file: 1_intro.tex
\section{Introduction}

\begin{figure}[t]
    \centering
    \begin{subfigure}[b]{0.49\linewidth}
        \includegraphics[width=\textwidth]{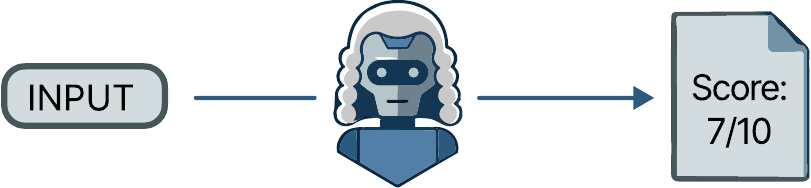}
        \caption{LLM-as-a-Judge}
        \label{subfig:llm-as-a-judge}
    \end{subfigure}
    \\
    \begin{subfigure}[b]{0.85\linewidth}
        \includegraphics[width=\textwidth]{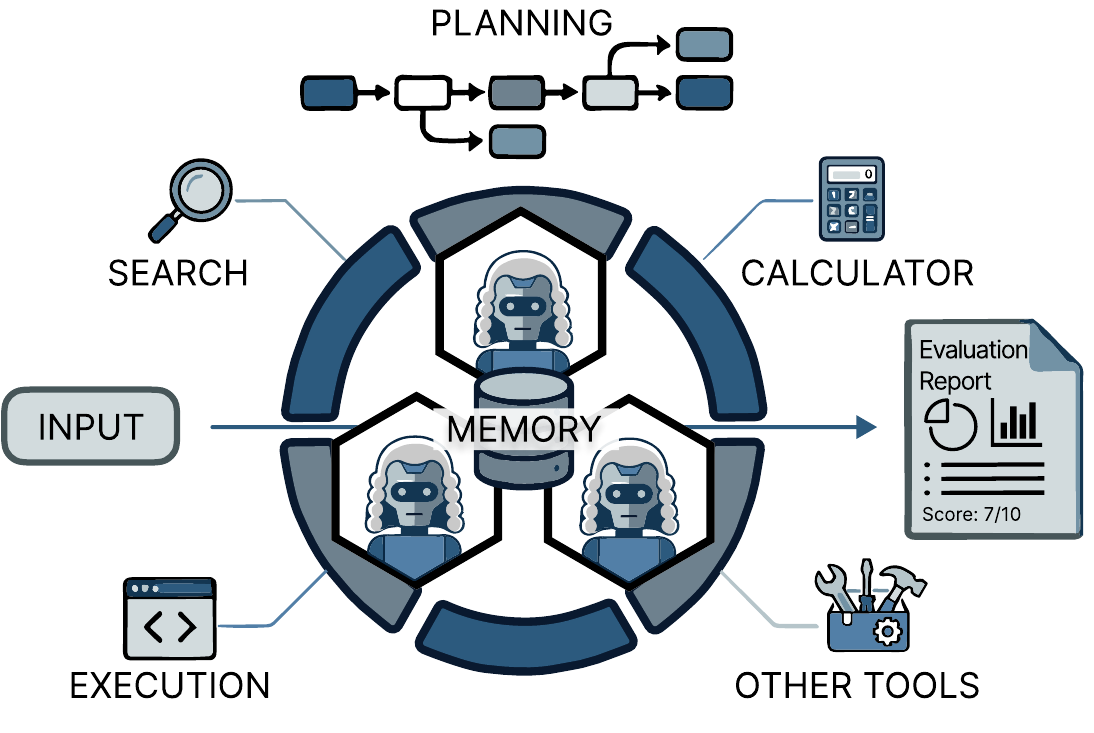}
        \caption{Agent-as-a-Judge}
        \label{subfig:agent-as-a-judge}
    \end{subfigure}
    \caption{Comparison between LLM-as-a-Judge (\ref{subfig:llm-as-a-judge}) and Agent-as-a-Judge (\ref{subfig:agent-as-a-judge}). The former performs direct single-pass evaluation, while the latter leverages planning, memory, and tool-augmented capabilities for enhanced evaluation.}
    \label{fig:paradigm-comparison}
\end{figure}

The rapid advancement of Large Language Models (LLMs) has revolutionized the field of AI evaluation, giving rise to the \textit{LLM-as-a-Judge} paradigm~\citep{Zheng2023Judging}.
While traditional metrics fail to capture semantic nuance and human judgment remains unscalable, this new approach leverages LLMs' advanced understanding and decision-making abilities to deliver near-human quality assessments across diverse domains~\citep{li2025LLM-as-a-Judge-Survey}.
Moreover, serving as a scalable proxy for human preference, LLM judges can provide reward signals for reinforcement learning~\citep{lee2024RLAIF-vs-RLHF} and enable the automated curation of massive synthetic datasets~\citep{long2024llmsdatageneration}. 
As such, LLM judgment has established itself as a cornerstone of AI evaluation and optimization pipelines, where the precision of the judge critically determines the success of downstream applications~\citep{lai2025post-training-scaling}.

However, as generative AI applications evolve from simple text responses to complex, multi-step tasks across specialized domains, the reliability of \textit{LLM-as-a-Judge} has become inevitably constrained~\cite{li2025LLM-as-a-Judge-Survey,xi2025LLM-agent-survey}.
First, single-pass evaluators are prone to inherent parametric biases—such as favoring verbosity or their own output patterns—which compromise their neutrality when assessing high-complexity responses that deviate from their training distribution~\cite{wang2024-LLMs-are-not-fair}.
Second, naive LLM judges are passive observers, unable to react to real-world observations; they assess answers based on linguistic patterns without verification, leading to hallucinated evaluations in specialized domains~\cite{peng2025AgenticRM}.
\
Furthermore, in evaluation tasks that require multifaceted assessment rubrics, traditional LLM judges experience cognitive overload when attempting to evaluate all dimensions comprehensively within a single inference step, which results in coarse-grained scores that fail to reflect specific nuances~\cite{zhang2025HiMATE}.

These limitations have catalyzed the transition from \textit{LLM-as-a-Judge} to \textit{Agent-as-a-Judge}.
\ 
As shown in Figure~\ref{fig:paradigm-comparison}, agentic judges proactively engage in evaluation through multiple capabilities: \
they decompose complex objectives into subtasks, mitigate biases through multi-agent collaboration~\cite{chan2024ChatEval},
ground assessments via tool-augmented evidence collection and correctness verification~\cite{peng2025AgenticRM},  
and enable fine-grained assessment by persisting intermediate states, autonomously planning the evaluation across reasoning steps~\cite{zhao2025RLPA,ding2025ARMThinker}. 
\
This paradigm shift enables more robust, verifiable, and nuanced assessments that effectively address the multifaceted nature of sophisticated AI-generated evaluands.

Despite the above potentials and rapid proliferation of agentic evaluation systems, the field lacks a survey to summarize and navigate this shifting landscape. To bridge this gap, we present the first comprehensive survey for \textit{Agent-as-a-Judge} through the following contributions:
\begin{itemize}
\setlength\itemsep{.05em}
    \item We identify and characterize the shift from \textit{LLM-as-a-Judge} to \textit{Agent-as-a-Judge} and summarize the agentic judges' development trend into three progressive stages (Section~\ref{sec:preliminary}).
    \item We organize core methodologies into five key parts according to agent's abilities (Section~\ref{sec:methods})  and survey their applications across general and professional domains (Section~\ref{sec:apps}).
    \item We analyze frontier challenges and identify promising research directions, providing a strategic roadmap for the next generation of robust and verifiable AI judgment.
\end{itemize}

%% file: 2_background.tex
\section{Evolution: From \textit{LLM-as-a-Judge} to \textit{Agent-as-a-Judge}}\label{sec:preliminary}
\input{forest-figure}

This section traces the evolution of automated evaluation from \textit{LLM-as-a-Judge} to \textit{Agent-as-a-Judge} paradigms. 
We begin by reviewing the foundational \textit{LLM-as-a-Judge} and its limitations.
We then examine the shift toward \textit{Agent-as-a-Judge}, analyzing key dimensions that characterize the agentic approach.
Finally, we summarize \textit{Agent-as-a-Judge}'s development trend into three progressive stages with distinct levels of autonomy and adaptability.

\subsection{\textit{LLM-as-a-Judge}}
\textit{LLM-as-a-Judge} paradigm emerged to overcome the scalability limits of human evaluation and the semantic insensitivity of traditional metrics. Zheng et al., \shortcite{Zheng2023Judging} formalized this approach by introducing benchmarks like MT-Bench to assess model alignment. Building on this, G-Eval \citep{liu2023G-Eval} leveraged chain-of-thought prompting for better alignment in natural language generation (NLG), while Prometheus \citep{kim2023Prometheus} induced fine-grained evaluation in open-source models via specialized tuning. To mitigate systematic issues like position and verbosity bias \citep{wang2024-LLMs-are-not-fair}, JudgeLM \citep{zhu2025judgelm} utilized fine-tuning to develop more robust evaluators.

\subsection{From \textit{LLM-as-a-Judge} to \textit{Agent-as-a-Judge}}
As evaluands evolve from simple text responses to complex, multi-step tasks across specialized domains, traditional \textit{LLM-as-a-Judge} has become increasingly inadequate, focusing on final outputs while failing to verify intermediate actions or satisfy the rigorous standards of professional fields~\cite{li2025LLM-as-a-Judge-Survey,xi2025LLM-agent-survey}. 
To bridge this gap, the paradigm is shifting toward \textit{Agent-as-a-Judge} that employs decentralized deliberation, executable verification, and fine-grained assessment to mitigate these limitations.

\paragraph{Evolving Robustness: From Monolithic to Decentralized.}
To mitigate the inherent parametric biases of monolithic LLM judges—such as the tendency to favor verbosity or their own output patterns—\textit{Agent-as-a-Judge} paradigms employ specialized, decentralized agents that collaborate through autonomous decision-making~\cite{chan2024ChatEval,chen2025MultiAgentasJudge}.
Crucially, this decentralized architecture facilitates the injection of expert prior knowledge: by decomposing complex evaluation goals into sub-tasks or structuring specific interaction workflows, we can enforce domain-specific constraints that a generalist model typically overlooks~\cite{xie2024grade,zhenxuan2025GEMA-Score}. Furthermore, multi-agent deliberation ensures collective robustness; distinct roles can isolate specific information points to neutralize bias, while debate and self-reflection allow agents to audit their own cognitive shortcuts, ensuring the final judgment transcends the heuristics of any single model~\cite{lyu2025debiasingLLM,wang2024-LLMs-are-not-fair}.

\paragraph{Evolving Verification: From Intuition to Execution.}
Static LLM judges are fundamentally passive observers, unable to react to real-world feedback. They assess answers based on linguistic plausibility -- how correct a response looks -- without verification or evidence collection, leading to "hallucinated correctness" in complex tasks~\cite{peng2025AgenticRM}.
\textit{Agent-as-a-Judge} bridges this reality gap by replacing intuition with execution. By interacting with external environments, agentic judges can query system states to verify side effects (e.g., file operations)~\cite{zhuge2025AgentasaJudge,wang2025CodeVisionary}, use code interpreters or theorem provers to validate logical consistency~\cite{ospanov2025HERMES}, and employ search tools to ground factual claims in real-time documentation~\cite{han2025VerifiAgent,peng2025AgenticRM}. This shifts the evaluative anchor from internal model knowledge to objective verification.

\paragraph{Evolving Granularity: From Global to Fine-grained.}
Many evaluation tasks inherently require multifaceted assessment rubrics, yet traditional LLM judges face a cognitive overload to evaluate these dimensions comprehensively within a single inference step, results in coarse-grained scores that fail to reflect specific nuances~\cite{zhang2025HiMATE}.
\textit{Agent-as-a-Judge} addresses this by transforming evaluation from a single-pass inference into autonomous, hierarchical reasoning~\cite{zhang2025HiMATE}. Instead of a monolithic assessment, an agentic judge can dynamically select or create task-specific rubrics, autonomously planning the evaluation to examine each component of the evaluand independently~\cite{ghosh2024SAGEval}, utilizing memory to track historical reasoning states and synthesize fragmented evidence into a coherent verdict. Consequently, these agents can pinpoint specific flaws that would otherwise be obscured in a global score, providing fine-grained feedback on each aspect~\cite{li2025AGENT-X}.

\subsection{\textit{Agent-as-a-Judge}}
\textit{Agent-as-a-Judge} represents a rapidly expanding field where the term "agent" is often applied loosely, spanning a heterogeneous range from procedural agentic workflows to autonomous self-evolvers~\citep{chan2024ChatEval,li2025AGENT-X,ding2025ARMThinker}. To provide a clear roadmap through this complexity, we summarize the ongoing development of agency as follows.

\paragraph{Procedural \textit{Agent-as-a-Judge}} decouples monolithic inference into agentic predefined workflows~\cite{su2025CAFES,zhenxuan2025GEMA-Score}
or engages in structured discussions among fixed sub-agents~\cite{chan2024ChatEval,feng2025MMAD}.
These systems enable complex judgments through coordinated multi-agent interactions, yet remain constrained by predetermined decision rules that cannot adapt to novel evaluation scenarios.

\paragraph{Reactive \textit{Agent-as-a-Judge}} enables adaptive decision-making by routing execution paths~\cite{zhang2025EvaluationAgent, li2025AGENT-X} and invoking external tools~\cite{peng2025AgenticRM} or sub-agents~\cite{chen2025MultiAgentasJudge} based on intermediate feedback. However, such reactivity remains confined to conditional routing within fixed decision spaces, lacking autonomy to refine underlying rubrics.

\paragraph{Self-Evolving \textit{Agent-as-a-Judge}} represents the cutting edge of the field, characterized by high autonomy and the ability to refine internal components during operation—synthesizing evaluation rubrics on-the-fly~\cite{wadhwa2025EvalAgents} and updating memory with lessons learned. This paradigm opens new frontiers for adaptive evaluation systems, though challenges remain in ensuring stability during self-modification~\cite{gao2025self-evolving-agent-survey}.

%% file: forest-figure.tex
    \pgfdeclarehorizontalshading{gradall}{100bp}{
        color(0bp)=(softsage);
        color(40bp)=(softsage);
        color(45bp)=(softsky);
        color(55bp)=(softsky);
        color(60bp)=(softpurple);
        color(100bp)=(softpurple)
    }
    \pgfdeclarehorizontalshading{grad23}{100bp}{
        color(0bp)=(softsky);
        color(35bp)=(softsky);
        color(55bp)=(softpurple);
        color(100bp)=(softpurple)
    }
    \pgfdeclarehorizontalshading{grad12}{100bp}{
        color(0bp)=(softsage);
        color(35bp)=(softsage);
        color(55bp)=(softsky);
        color(100bp)=(softsky)
    }
    
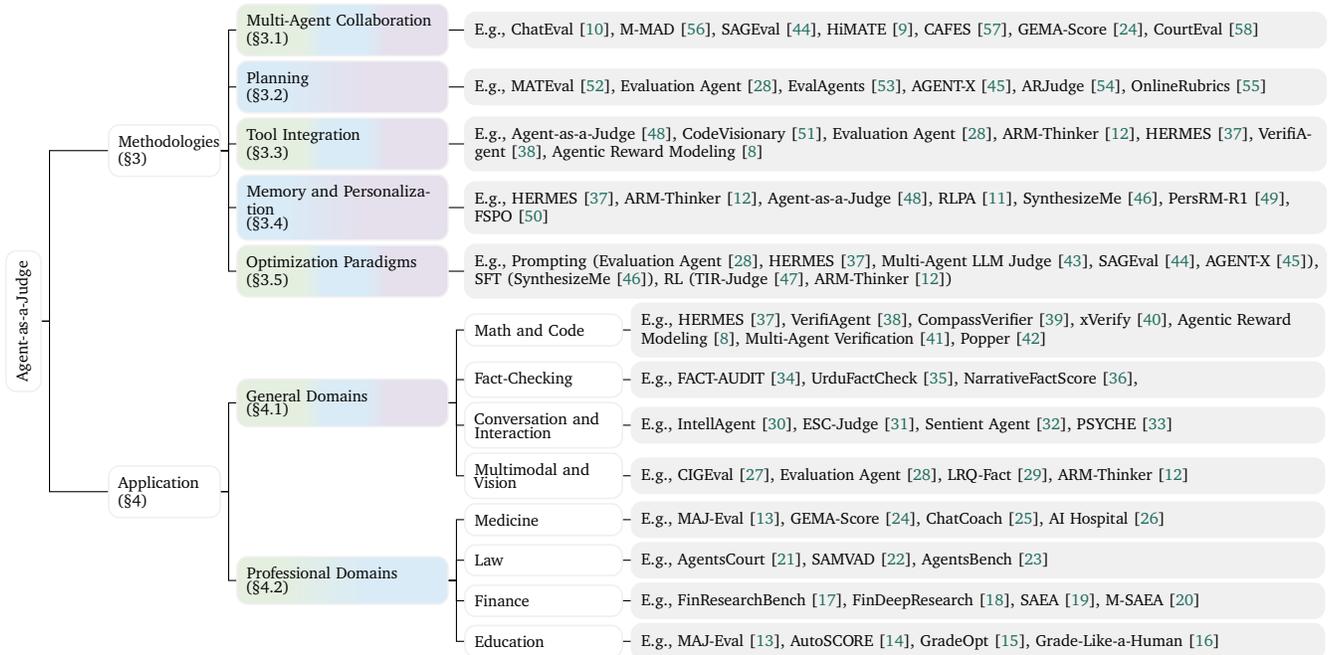
\begin{figure*}[tbhp]

    \pgfkeys{/forest/col1/.style={text width=1.2cm}}
    \pgfkeys{/forest/col2/.style={text width=2.5cm}}
    \pgfkeys{/forest/col3/.style={text width=1.8cm}}
    \pgfkeys{/forest/colContent/.style={text width=8.8cm}}
    \pgfkeys{/forest/colWideContent/.style={text width=11cm}}
    \pgfkeys{/forest/gradallContent/.style={
        shading=gradall, shading angle=0
    }}
    \pgfkeys{/forest/grad23Content/.style={
        shading=grad23, shading angle=0
    }}
    \pgfkeys{/forest/grad12Content/.style={
        shading=grad12, shading angle=0
    }}
    \resizebox{\textwidth}{!}{
    \begin{forest}
    for tree={   
        font=\fontsize{6}{5}\selectfont,
        draw=softgray, semithick, rounded corners,
        minimum height = 1.ex,
        minimum width = 2em,
        anchor = west,
        grow = east,
        forked edge,        
        s sep = 0.5mm,    
        l sep = 2mm,    
        fork sep = 1mm,            
    }
    [Agent-as-a-Judge, anchor=center, rotate=90
        [Application (\S \ref{sec:apps}), fit=band, col1
            [Professional Domains \\(\S \ref{sec:apps:professional}), col2, grad12Content
                [Education, col3, l sep = 1mm
                    [{E.g., MAJ-Eval~\shortcite{chen2025MultiAgentasJudge},
                    AutoSCORE~\shortcite{wang2025autoscore},
                    GradeOpt~\shortcite{chu2024llm},
                    Grade-Like-a-Human~\shortcite{xie2024grade}
                    }, colContent, fill=softgray]
                ]
                [Finance, col3, l sep = 1mm
                    [{E.g.,
                    FinResearchBench~\shortcite{sun2025finresearchbench},
                    FinDeepResearch~\shortcite{zhu2025findeepresearch},
                    SAEA~\shortcite{chen2025standardbenchmarksfail},
                    M-SAEA~\shortcite{chen2025tasks}
                    }, colContent, fill=softgray]
                ]
                [Law, col3, l sep = 1mm
                    [{E.g., AgentsCourt~\shortcite{he2024agentscourt}, 
                    SAMVAD~\shortcite{devadiga2025samvad},
                    AgentsBench~\shortcite{jiang2025agentsbench}
                    }, colContent, fill=softgray]
                ]
                [Medicine, col3, l sep = 1mm
                    [{E.g., MAJ-Eval~\shortcite{chen2025MultiAgentasJudge},
                    GEMA-Score~\shortcite{zhenxuan2025GEMA-Score},
                    ChatCoach~\shortcite{huang2024benchmarking},
                    AI Hospital~\shortcite{fan2025ai}
                    }, colContent, fill=softgray]
                ]
            ]
            [General Domains \\(\S \ref{sec:apps:general}), col2, gradallContent
                [Multimodal and Vision, col3, l sep = 1mm
                    [{E.g., CIGEval~\shortcite{wang-etal-2025-unified}, 
                    Evaluation Agent~\shortcite{zhang2025EvaluationAgent}, 
                    LRQ-Fact~\shortcite{beigi2024lrqfact}, 
                    ARM-Thinker~\shortcite{ding2025ARMThinker}}, colContent, fill=softgray]
                ]
                [Conversation and Interaction, col3, l sep = 1mm
                    [{E.g., IntellAgent~\shortcite{levi2025intellagent}, 
                    ESC-Judge~\shortcite{madani-srihari-2025-esc}, 
                    Sentient Agent~\shortcite{zhang2025sentient}, 
                    PSYCHE~\shortcite{lee2025psyche}
                    }, colContent, fill=softgray]
                ]
                [Fact-Checking, col3, l sep = 1mm
                    [{E.g., FACT-AUDIT~\shortcite{lin-etal-2025-fact}, 
                    UrduFactCheck~\shortcite{ahmad-etal-2025-urdufactcheck}, 
                    NarrativeFactScore~\shortcite{jeong-etal-2025-agent}, 
                    }, colContent, fill=softgray]
                ]
                [Math and Code, col3, l sep = 1mm
                    [{E.g., HERMES~\shortcite{ospanov2025HERMES}, 
                    VerifiAgent~\shortcite{han2025VerifiAgent}, 
                    CompassVerifier~\shortcite{liu-etal-2025-compassverifier}, 
                    xVerify~\shortcite{chen2025xverify}, 
                    Agentic Reward Modeling~\shortcite{peng2025AgenticRM},
                    Multi-Agent Verification~\shortcite{lifshitz2025multi},
                    Popper~\shortcite{huang2025popper}
                    }, colContent, fill=softgray]
                ]
            ]
        ]
        [Methodologies \\(\S \ref{sec:methods}), fit=band, col1
            [Optimization Paradigms \\(\S \ref{sec:optimization}), col2, gradallContent, l sep = 2mm
                [{E.g., Prompting (Evaluation Agent~\shortcite{zhang2025EvaluationAgent}, HERMES~\shortcite{ospanov2025HERMES}, Multi-Agent LLM Judge~\shortcite{cao2025MultiAgentLLMJudge}, SAGEval~\shortcite{ghosh2024SAGEval}, AGENT-X~\shortcite{li2025AGENT-X}), SFT (SynthesizeMe~\shortcite{ryan2025SynthesizeMe}), RL (TIR-Judge~\shortcite{xu2025TIRJudge}, ARM-Thinker~\shortcite{ding2025ARMThinker})}, colWideContent, fill=softgray]
            ]
            [Memory and Personalization \\(\S \ref{sec:memory}), col2, grad23Content, l sep = 2mm
                [{E.g., HERMES~\shortcite{ospanov2025HERMES},
                ARM-Thinker~\shortcite{ding2025ARMThinker}, 
                Agent-as-a-Judge~\shortcite{zhuge2025AgentasaJudge},
                RLPA~\shortcite{zhao2025RLPA},
                SynthesizeMe~\shortcite{ryan2025SynthesizeMe},
                PersRM-R1~\shortcite{li2025PersRMR1},
                FSPO~\shortcite{singh2025FSPO}
                }, colWideContent, fill=softgray]
            ]
            [Tool Integration \\(\S \ref{sec:tool}), col2, gradallContent, l sep = 2mm
                [{E.g., Agent-as-a-Judge~\shortcite{zhuge2025AgentasaJudge},
                CodeVisionary~\shortcite{wang2025CodeVisionary},
                Evaluation Agent~\shortcite{zhang2025EvaluationAgent},
                ARM-Thinker~\shortcite{ding2025ARMThinker},
                HERMES~\shortcite{ospanov2025HERMES},
                VerifiAgent~\shortcite{han2025VerifiAgent}, 
                Agentic Reward Modeling~\shortcite{peng2025AgenticRM}
                }, colWideContent, fill=softgray]
            ]
            [Planning \\(\S \ref{sec:plan}), col2, grad23Content, l sep = 2mm
                [{E.g.,
                MATEval~\shortcite{Li2024MATEval},
                Evaluation Agent~\shortcite{zhang2025EvaluationAgent},
                EvalAgents~\shortcite{wadhwa2025EvalAgents},
                AGENT-X~\shortcite{li2025AGENT-X},
                ARJudge~\shortcite{xu2025ARJudge},
                OnlineRubrics~\shortcite{rezaei2025OnlineRubrics}
                }, 
                colWideContent,
                fill=softgray]
            ]
            [Multi-Agent Collaboration \\(\S \ref{sec:multiagent}), col2, gradallContent, l sep = 2mm
                [{E.g.,
                ChatEval~\shortcite{chan2024ChatEval},
                M-MAD~\shortcite{feng2025MMAD},
                SAGEval~\shortcite{ghosh2024SAGEval},
                HiMATE~\shortcite{zhang2025HiMATE},
                CAFES~\shortcite{su2025CAFES},
                GEMA-Score~\shortcite{zhenxuan2025GEMA-Score},
                CourtEval~\shortcite{kumar2025courteval}
                }, 
                colWideContent,
                fill=softgray]
            ]
        ]
    ]
    \end{forest}
    }
    \caption{A taxonomy of Agent-as-a-Judge organizing Methodologies (\S \ref{sec:methods}) and Applications (\S \ref{sec:apps}). Background gradients illustrate the coverage of developmental stages, from \colorbox{softsage}{\textit{Procedural}} to \colorbox{softsky}{\textit{Reactive}} and then to \colorbox{softpurple}{\textit{Self-Evolving}}.}
    \label{fig:taxonomy}
    \end{figure*}

%% file: 3_taxonomy.tex
\section{Methodologies}\label{sec:methods}

This section categorizes \textit{Agent-as-a-Judge} methodologies into five dimensions: multi-agent collaboration, planning, tool integration, memory and personalization, and optimization paradigms.
As shown in Figure~\ref{fig:taxonomy}, implementation sophistication reveals the evolutionary stages: foundational methodologies (collaboration, tool integration, optimization) evolve across all stages, while others (planning, memory) emerge more prominently in advanced paradigms.
The following subsections examine how each methodology manifests across these stages.

\subsection{Multi-Agent Collaboration}
\label{sec:multiagent}
\begin{figure}[t]
    \centering
    \begin{subfigure}[b]{0.4\linewidth}
        \includegraphics[width=\textwidth]{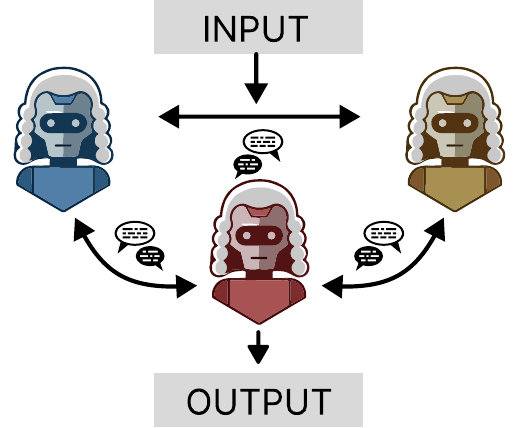}
        \caption{Collective Consensus}
        \label{subfig:multiagent1}
    \end{subfigure}
    \hfill
    \begin{subfigure}[b]{0.4\linewidth}
        \includegraphics[width=\textwidth]{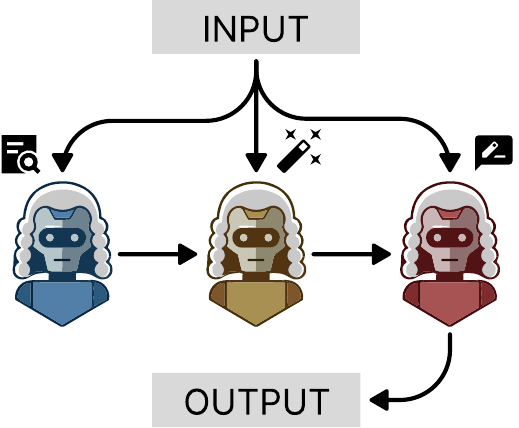}
        \caption{Task Decomposition}
        \label{subfig:multiagent2}
    \end{subfigure}
    \caption{Multi-agent collaboration paradigms.}
    \label{fig:multiagent-topologies}
\end{figure}

Multi-agent collaboration leverages collective reasoning to mitigate single-LLM biases in \textit{Agent-as-a-Judge} systems.
Early systems followed \textit{Procedural} paradigms with fixed protocols, while recent work evolves toward \textit{Reactive} approaches that adapt agent selection based on feedback.
We categorize these into two topologies:

\paragraph{Collective Consensus.}
Horizontal debate mechanisms leverage agents representing diverse perspectives to counteract the inherent biases of single-LLM evaluators, illustrated in Figure~\ref{fig:multiagent-topologies}.
Early approaches exemplified the \textit{Procedural} stage: ChatEval~\cite{chan2024ChatEval} pioneered this with a courtroom-inspired discussion mechanism where agents debate as equals following predefined protocols. This paradigm was later extended to machine translation in M-MAD~\cite{feng2025MMAD}, while subsequent research~\cite{koupaee2025Faithful} introduced explicit stances and "judge" roles to prevent agents from blindly conforming to the majority.
Recent methods have become more \textit{Self-Evolving}: approaches like Multi-agent-as-judge~\cite{chen2025MultiAgentasJudge} have moved beyond static ensembles by creating domain-specific experts based on intermediate feedback.

\paragraph{Task Decomposition.}
Task Decomposition employs a "Divide and Conquer" strategy, delegating distinct subtasks to specialized agents for systematic evaluation, illustrated in Figure~\ref{fig:multiagent-topologies}.
Early frameworks followed \textit{Procedural} designs: sequential approaches like CAFES~\cite{su2025CAFES} and GEMA-Score~\cite{zhenxuan2025GEMA-Score,kumar2025courteval} structure evaluation into predefined stages (e.g., Evidence Gathering, Reasoning, Scoring), while SAGEval~\cite{ghosh2024SAGEval} introduces supervision via a "Judge the Judge" meta-evaluator that reviews previous agents' decisions, with hierarchical approaches like HiMATE~\cite{zhang2025HiMATE} organizing agents into tree structures for varying error granularities.
More recent work has shifted toward \textit{Reactive} paradigms: AGENT-X~\cite{li2025AGENT-X} employs adaptive router agent that dynamically selects the most relevant base agents based on intermediate analysis results.

\begin{takeaway}{Takeaway}
    {
Multi-agent evaluation frameworks adopt two main topologies: Collective Consensus and Task Decomposition. Recent advances have evolved toward more autonomous systems that can select or generate subagents.
    }
\end{takeaway}

\subsection{Planning}
\label{sec:plan}
\revise{todo: a figure}
Planning serves as a core capability in the \textit{Agent-as-a-Judge} paradigm, enabling the decomposition of high-level evaluation objectives into executable sub-tasks and the dynamic adaptation of assessment trajectories based on intermediate analysis.
This section examines planning capabilities from two perspectives:

\paragraph{Workflow Orchestration.}
Workflow orchestration in \textit{Agent-as-a-Judge} systems spans from static frameworks to dynamic agency, primarily characterizing \textit{Procedural} and \textit{Reactive} stages of agentic evaluation.
Approaches like MATEval~\cite{Li2024MATEval} rely on static decomposition, breaking tasks into fixed sequences of sub-dimensions. While this ensures systematic assessment through predefined control flows, it limits adaptability in complex scenarios.
Conversely, Evaluation Agent~\cite{zhang2025EvaluationAgent} introduces dynamic multi-round planning, where agents adjust strategies based on intermediate feedback. This system further optimizes efficiency through autonomous termination, allowing the agent to self-monitor information gain and proactively halt execution once sufficient evidence is gathered.

\paragraph{Rubric Discovery.}
Unlike general agents focused on task completion, Judge Agents have the distinct capability to autonomously formulate and refine rubrics, representing a hallmark of the \textit{Self-Evolving} stage, where agents can refine their internal evaluation components.
EvalAgents~\cite{wadhwa2025EvalAgents} exemplifies this by employing a Query Generator that plans web searches to discover implicit rubrics, while AGENT-X~\cite{li2025AGENT-X} uses an Adaptive Router to infer domain context and plan bespoke detection guidelines.
ARJudge~\cite{xu2025ARJudge} adaptively formulates rubrics by iteratively generating context-sensitive questions, and OnlineRubrics~\cite{rezaei2025OnlineRubrics} integrates planning into reinforcement learning, evolving rubrics alongside policy optimization to detect reward hacking.

\begin{takeaway}
    {Takeaway}
    {Serving as the strategic engine, planning shifts evaluation from rigid flows to adaptive exploration, enabling agents to optimize \textit{how} they evaluate (workflow orchestration) and \textit{what} they evaluate (rubric discovery).
    }
    
\end{takeaway}

\subsection{Tool Integration} 
\label{sec:tool}











\begin{table*}[t]
\centering
\small
\resizebox{1\textwidth}{!}{
\begin{tabularx}{\textwidth}{C{2.2cm} l l X}
\toprule
\textbf{Tool Purpose} & \textbf{Method} & \textbf{Evaluation Task} & \textbf{Tool Type} \\
\midrule

\multirow{4}{2cm}{Evidence collection} 
& Agent-as-a-Judge~\shortcite{zhuge2025AgentasaJudge} 
& Code generation 
& Graph, locate, read, search, retrieve \\

& CodeVisionary~\shortcite{wang2025CodeVisionary} 
& Code generation 
& Code execution, static linter, unit tests, screenshot, web browsing \\

& Evaluation Agent~\shortcite{zhang2025EvaluationAgent} 
& Visual generation 
& Visual generative models \\

& ARM-Thinker~\shortcite{ding2025ARMThinker} 
& Multimodal generation 
& Instruction following checks, crop/zoom tools, document retrieval tools\\

\midrule

\multirow{3}{2cm}{Correctness verification}
& HERMES~\shortcite{ospanov2025HERMES} 
& Math reasoning 
& Translator, theorem prover \\

& VerifiAgent~\shortcite{han2025VerifiAgent} 
& Factual \& Math reasoning 
& Search engine, Python interpreter, theorem prover \\

& Agentic RM~\shortcite{peng2025AgenticRM} 
& Factual \& Math reasoning 
& Search engine, Python interpreter \\

\bottomrule
\end{tabularx}
}
\caption{Tool integration in representative \textit{Agent-as-a-Judge} methods, grouped by primary tool usage purpose.}
\label{tab:tool}
\vspace{-1.5em}
\end{table*}

Tool integration is a defining capability of \textit{Agent-as-a-Judge} frameworks, enabling judges to ground evaluation in external evidence and explicit checks. As shown in Table~\ref{tab:tool}, existing approaches can be grouped into evidence collection and correctness verification based on the purpose of tool use.

\paragraph{Evidence Collection.}
A common use of tools in \textit{Agent-as-a-Judge} frameworks is to collect additional evidence that supports evaluations. Such evidence includes intermediate artifacts, execution results, and perceptual signals that cannot be reliably obtained through text-based reasoning. In code-related tasks, Agent-as-a-Judge~\cite{zhuge2025AgentasaJudge} and CodeVisionary~\cite{wang2025CodeVisionary} allow judges to inspect execution artifacts or run automated checks to expose execution feedback for evaluation. Similar methods are adopted in multimodal settings. Evaluation Agent~\cite{zhang2025EvaluationAgent} enables judges to invoke external visual models to obtain visual quality or alignment signals, while ARM-Thinker~\cite{ding2025ARMThinker} gathers fine-grained visual and contextual evidence through document access and localized visual operations. Overall, these works integrate tools to surface observable and task-relevant evidence, expanding the judge’s access to execution-level, perceptual, and contextual information, and supporting more reliable evaluation.

\paragraph{Correctness Verification.}
Another line of work employs tools to verify whether the evaluand’s outputs or intermediate reasoning steps satisfy explicit correctness constraints, such as logical validity, mathematical soundness, or factual consistency. In these frameworks, the judge agent identifies which claims or steps require verification and invokes appropriate tools to check them. The resulting verification signals are then interpreted by the agent in context to inform the final evaluation. HERMES~\cite{ospanov2025HERMES} verifies mathematical reasoning through formal theorem proving, while VerifiAgent~\cite{han2025VerifiAgent} invokes programmatic and symbolic checkers to validate factual and computational claims. Agentic Reward Modeling~\cite{peng2025AgenticRM} further integrates correctness verification by combining fact-checking tools and programmatic validators to produce structured correctness signals that inform the final evaluation.

\begin{takeaway}
    {Takeaway}
    {
        Tool integration in \textit{Agent-as-a-Judge} grounds evaluation in observable and verifiable signals by allowing judges to actively gather evidence and check correctness through external tools.
    }
\end{takeaway}

\subsection{Memory and Personalization}
\label{sec:memory}
Memory enables \textit{Agent-as-a-Judge} frameworks to retain information across evaluation steps, supporting multi-step reasoning, consistent judgment, and reuse of prior results. We categorize prior work by the role of memory, including intermediate state tracking and personalized context preservation.

\paragraph{Intermediate State.}
In multi-step evaluation settings, \textit{Agent-as-a-Judge} frameworks use memory to retain intermediate evaluation states generated during assessment, providing the necessary context for conditional routing and adaptive decision-making based on intermediate feedback--a fundamental mechanism for \textit{Reactive Agent-as-a-Judge}.
HERMES~\cite{ospanov2025HERMES} retains intermediate proof states when combining reasoning with formal theorem proving, enabling consistent verification across long reasoning chains. ARM-Thinker~\cite{ding2025ARMThinker} preserves intermediate evidence such as visual reasoning outputs and tool interaction results, which are later reused to ground evaluation. Agent-as-a-Judge~\cite{zhuge2025AgentasaJudge} records execution traces and step-level feedback, enabling evaluation beyond final outputs to account for intermediate behavior. Collectively, these methods use memory to retain intermediate states that support cumulative, step-aware evaluation.

\paragraph{Personalized Context.}
\textit{Agent-as-a-Judge} frameworks often incorporate memory to retain user-related information that conditions evaluation across interactions. Such memory captures user preferences, evaluation standards, or prior feedback, allowing judgments to remain consistent over time. PersRM-R1~\cite{li2025PersRMR1} and FSPO~\cite{singh2025FSPO} store preference data derived from historical interactions, including preference labels or few-shot examples, which are reused to condition subsequent evaluations for the same user. 
More advanced approaches abstract historical preference signals into persistent user personas or long-term profiles. RLPA~\cite{zhao2025RLPA} and SynthesizeMe~\cite{ryan2025SynthesizeMe} exemplify this by constructing and maintaining user personas that are stored and reused to guide evaluation. 
Such long-term user profiling often serves to support \textit{Self-Evolving Agent-as-a-Judge}, enabling continuous optimization based on evolving preferences. Together, these methods use memory to preserve personalized context that shapes evaluation behavior and ensures consistency across interactions.

\begin{takeaway}
    {Takeaway}
    {
        Memory enables \textit{Agent-as-a-Judge} to preserve intermediate states and personalized context, supporting multi-step evaluation, consistent judgment, and efficient reuse of prior information.
    }
\end{takeaway}

\subsection{Optimization Paradigms}
\label{sec:optimization}
Optimization paradigms define how \textit{Agent-as-a-Judge} improves evaluation quality by updating model parameters or adapting evaluation behaviors. We organize prior work into two groups: training-time optimization and inference-time optimization.

\paragraph{Training-Time Optimization.}
Training-time optimization improves \textit{Agent-as-a-Judge} by updating model parameters to align judgment behavior with evaluation objectives. Supervised fine-tuning is commonly used to standardize judge behavior, training models to follow explicit criteria, and produce structured judgments across tasks. For example, SynthesizeMe~\cite{ryan2025SynthesizeMe} shapes evaluation behavior using persona-guided supervision derived from historical data. Reinforcement learning optimizes judges to perform evaluation actions more effectively, especially in settings that require tool use and multi-step verification. TIR-Judge~\cite{xu2025TIRJudge} and ARM-Thinker~\cite{ding2025ARMThinker} train judges to decide when and how to invoke tools, integrate external signals, and verify intermediate results. Together, training-time optimization shapes internal decision processes, enabling more reliable, structured evaluation.

\paragraph{Inference-Time Optimization.}
Inference-time optimization adapts evaluation behavior without updating model parameters by controlling how judgments are produced through prompts, workflows, or agent interactions. Existing approaches can be broadly grouped into two types. 1) The first type follows predefined evaluation procedures, where reasoning steps, verification routines, or prompts are fixed in advance to ensure consistency and efficiency. Evaluation Agent~\cite{zhang2025EvaluationAgent} and HERMES~\cite{ospanov2025HERMES} exemplify this setting by adopting structured, step-by-step evaluation pipelines. 2) The second type allows evaluation behavior to adapt during inference, where the evaluation process, participating agents, or applied criteria change based on intermediate results. Multi-Agent LLM Judge~\cite{cao2025MultiAgentLLMJudge} iteratively refines prompts and context through multi-judge coordination, while SAGEval~\cite{ghosh2024SAGEval} introduces a meta-judge to monitor and revise judge behavior. ChatEval~\cite{chan2024ChatEval} and AGENT-X~\cite{li2025AGENT-X} further support adaptive evaluation through agent interaction and dynamic guideline selection. Overall, inference-time optimization enables flexible control over evaluation behavior, ranging from fixed procedures to adaptive, interaction-driven judgment.

\begin{takeaway}
    {Takeaway}
    {
        Optimization improves \textit{Agent-as-a-Judge} by either learning evaluation behavior through parameter updates at training-time or adjusting evaluation strategies at inference time.
    }
\end{takeaway}

%% file: 4_apps.tex
\section{Application}\label{sec:apps}
\begin{figure}[t]
\centering
\includegraphics[width=1\linewidth]{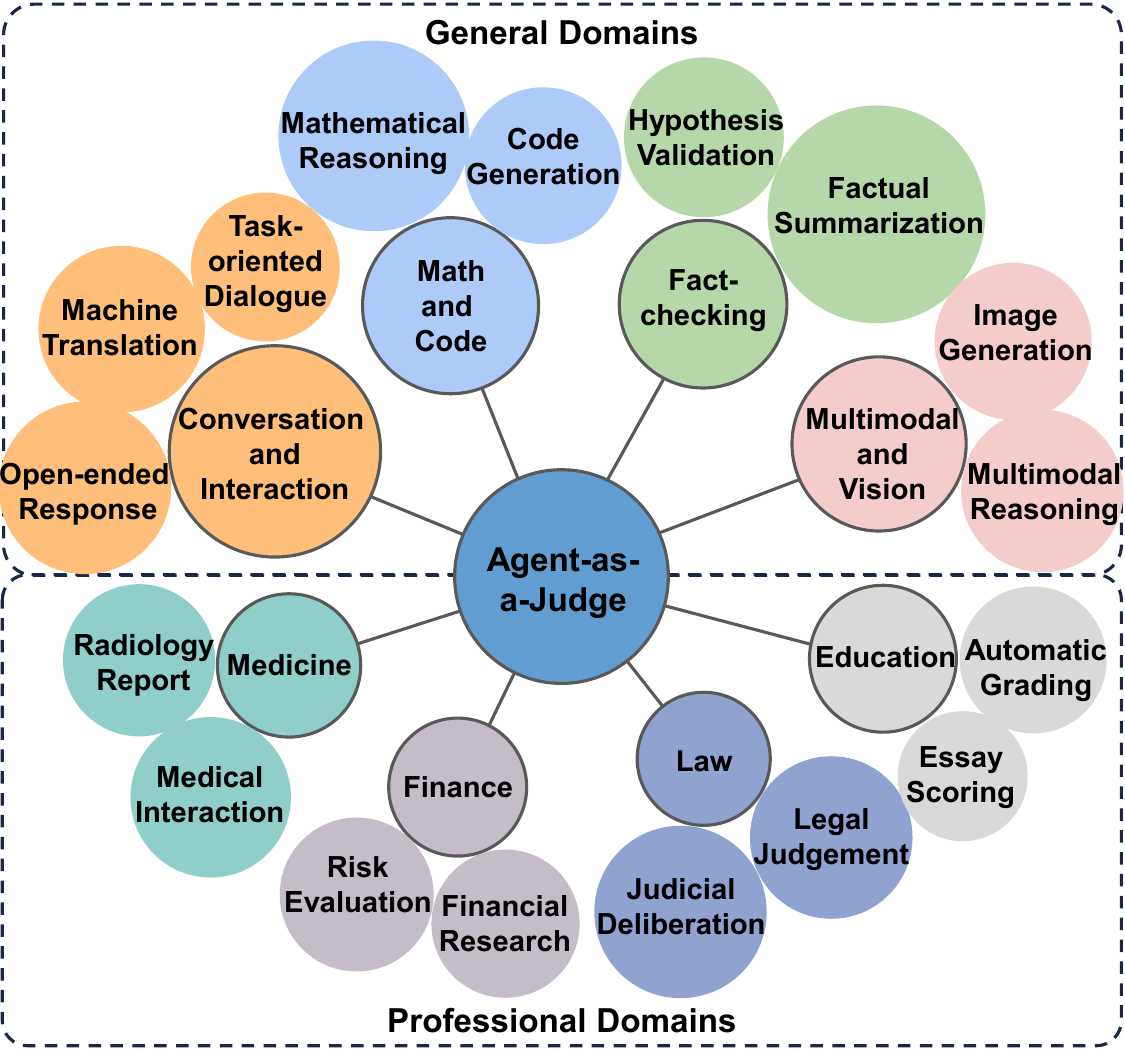}
\caption{An overview of \textit{Agent-as-a-Judge} application domains and their fine-grained task categories.}
\vspace{-1em}
\label{fig:app}
\end{figure}

Building on the methods above, this section describes how \textit{Agent-as-a-Judge} methods are applied in different evaluation tasks. As shown in Figure~\ref{fig:app}, we organize representative applications into two groups: general domains and professional domains.

\subsection{General Domains}\label{sec:apps:general}
\paragraph{Math and Code.}
In math and code evaluation, \textit{Agent-as-a-Judge} systems move beyond single-pass scoring by grounding judgment in verifiable reasoning signals. One line of work augments free-form reasoning with explicit correctness checks. HERMES~\cite{ospanov2025HERMES} anchors LLM reasoning to intermediate formal proof steps, reducing drift in long derivations. VerifiAgent~\cite{han2025VerifiAgent} decouples high-level reasoning assessment from tool-based correctness verification, enabling adaptive checking across reasoning types. CompassVerifier~\cite{liu-etal-2025-compassverifier} and xVerify~\cite{chen2025xverify} focus on mathematical and logical outputs, addressing equivalence checking under diverse surface forms. Other approaches strengthen judgment by aggregating multiple evaluation signals. Multi-Agent Verification~\cite{lifshitz2025multi} distributes evaluation across aspect-specific judges. Agentic Reward Modeling~\cite{peng2025AgenticRM} integrates preference-based supervision with verifiable correctness signals to improve reward reliability. Popper~\cite{huang2025popper} formulates judgment as controlled falsification, using statistical tests to validate free-form claims.

\paragraph{Fact-Checking.}
In fact-checking, \textit{Agent-as-a-Judge} reframes evaluation from static label prediction to interactive verification with evidence gathering and justification. FACT-AUDIT~\cite{lin-etal-2025-fact} models fact-checking as an agentic loop with multi-agent collaboration, jointly evaluating verdict accuracy and justification quality. This paradigm is particularly effective when evidence is scarce or inconsistencies are subtle. UrduFactCheck~\cite{ahmad-etal-2025-urdufactcheck} improves robustness in low-resource settings through multilingual retrieval and evidence boosting. NarrativeFastScore~\cite{jeong-etal-2025-agent} addresses long-context factual consistency by constructing character-level knowledge representations, enabling detection of state and relation errors with actionable feedback.

\paragraph{Conversation and Interaction.}
In conversation and interaction, \textit{Agent-as-a-Judge} shifts from grading isolated replies to constructing multi-turn exchanges, enabling 
evaluation under evolving goals, constraints, and 
user reactions. 
For task-oriented dialogue, IntellAgent~\cite{levi2025intellagent} uses interactive user simulations to synthesize conversational benchmarks, while \citet{kazi2024large} introduces frameworks for controllable user goals and automatic measures.
For affective and social interaction, ESC-Judge~\cite{madani-srihari-2025-esc} constructs emotional-support agents via standardized counseling skills, Sentient Agent~\cite{zhang2025sentient} tracks emotion trajectories over time to reflect higher-order 
social cognition, and PSYCHE~\cite{lee2025psyche} builds psychiatric patient profiles for ethical assessment validation.
\citet{wu2023large} frames evaluation as multi-perspective role play with diverse reviewer personas to cover both objective and 
subjective dimensions.

\paragraph{Multimodal and Vision.}
In the multimodal and vision domain, \textit{Agent-as-a-Judge} shifts from static scoring to interactive inspection.
For visual generation, CIGEval~\cite{wang-etal-2025-unified} orchestrates specialized tools to probe control adherence and subject consistency, while Evaluation Agent~\cite{zhang2025EvaluationAgent} runs multi-round checks to provide user-tailored, explainable analyses.
For truthfulness evaluation, LRQ-Fact~\cite{beigi2024lrqfact} generates targeted fact-checking questions across image and text to guide evidence retrieval, while ARM-Thinker~\cite{ding2025ARMThinker} selectively invokes tools like image inspection for finalizing judgments.

\subsection{Professional Domains} \label{sec:apps:professional}

\paragraph{Medicine.}
In high-stakes clinical NLP, \textit{Agent-as-a-Judge} appears in two forms: 1) multi-agent evaluators that decompose clinical quality into specialized roles, and 2) agentic simulators that interactively elicit clinical behaviors. 
For 1), MAJ-Eval~\cite{chen2025MultiAgentasJudge} constructs multiple evaluator personas to debate and cross-verify responses, while GEMA-Score\cite{zhenxuan2025GEMA-Score} uses agent collaboration to compute granular, tool-assisted scores covering disease severity and uncertainty. 
For 2), Chat-Coach~\cite{huang2024benchmarking} pairs autonomous \textit{patient} and \textit{coach} agents to critique trainee-doctor dialogues, while AI Hospital~\cite{fan2025ai} evaluates LLM ``doctors'' in multi-agent simulators, though final scoring often still requires conventional metrics.

\paragraph{Law.}
In the legal domain, \textit{Agent-as-a-Judge} simulates the adversarial and deliberative nature of jurisprudence through multi-agent interaction. AgentsCourt~\cite{he2024agentscourt} introduces adversarial debate frameworks where agents role-play as prosecutors, defense attorneys, and judges, exposing the evaluating agent to conflicting arguments to improve verdict robustness. SAMVAD~\cite{devadiga2025samvad} and AgentsBench~\cite{jiang2025agentsbench} model judicial consensus by simulating bench deliberation processes, capturing interactions between concurring and dissenting opinions to enhance legal judgment prediction.

\paragraph{Finance.}
In finance, \textit{Agent-as-a-Judge} addresses two limitations of static benchmarks: 1) capturing the \emph{internal research logic} of long-form analyst reports, and 2) detecting \emph{deployment risks} like hallucinations and temporal staleness. 
For 1), FinResearchBench~\cite{sun2025finresearchbench} extracts logic trees from reports as intermediate structures for comprehensive assessment, whereas FinDeepResearch~\cite{zhu2025findeepresearch} can synthesize hierarchical rubrics but still relies on predefined workflows.
For 2), SAEA~\cite{chen2025standardbenchmarksfail} proposes auditing agent \emph{trajectories} to mitigate hallucinations and temporal misalignment. From Tasks to Teams~\cite{chen2025tasks} extends this approach with M-SAEA to trace multi-agent failures, such as cross-agent divergence and error propagation.

\paragraph{Education.}
In the educational domain, \textit{Agent-as-a-Judge} systems emulate pedagogical nuance through collaborative, role-specialized workflows. Grade-Like-Human~\cite{xie2024grade} and AutoSCORE~\cite{wang2025autoscore} decompose grading into staged processes (rubric construction, evidence recognition, cross-review) to improve grounding and consistency. Beyond static scoring, MAJ-Eval~\cite{chen2025MultiAgentasJudge} uses multi-persona debates to align with multi-dimensional human evaluation, while GradeOpt~\cite{chu2024llm} introduces agents that diagnose discrepancies and iteratively refine grading guidelines.

%% file: 5_discussion.tex
\section{Discussion}\label{sec:discussion}
This section discusses broader issues that arise when deploying \textit{Agent-as-a-Judge} systems in practice. We first summarize key challenges that limit scalability, reliability, and real-world adoption, and then outline several future directions that may help address these limitations and further advance agentic evaluation.

\subsection{Challenges}
\textit{Agent-as-a-Judge} improves evaluation reliability through planning, tool use, memory, and multi-agent collaboration, but these capabilities also introduce new challenges beyond static \textit{LLM-as-a-Judge}. Key challenges include computational cost, latency, safety, and privacy.

\paragraph{Computational Cost.}
\textit{Agent-as-a-Judge} introduces a heavier computational burden in both training and inference. 1) Training a judge agent is expensive. Supervised fine-tuning alone is often insufficient to support agentic behaviors such as tool invocation, long-horizon planning, and adaptive decision making. Reinforcement learning provides a natural way to acquire these capabilities, but it significantly increases training cost, especially when the judge operates over long trajectories or complex tool-calling sequences. 2) Inference with \textit{Agent-as-a-Judge} is also costly. Unlike single-pass judgment, agentic evaluation typically involves multiple reasoning steps, intermediate decisions, and coordination among multiple agents, all of which increase computation per evaluation.

\paragraph{Latency.}
In addition to higher computational cost, \textit{Agent-as-a-Judge} often suffers from increased inference latency. Agentic evaluation requires sequential reasoning steps, external tool calls, or multi-agent communication, each of which introduces additional delays. This latency can be particularly problematic in real-time or interactive settings, such as online model evaluation, user-facing content moderation, or reinforcement learning loops where rapid feedback is required. As a result, there exists a tension between evaluation reliability and practical deployment constraints, where more thorough agentic judgment may not be feasible under strict latency budgets.

\paragraph{Safety.}
While \textit{Agent-as-a-Judge} is designed to improve evaluation robustness, it also raises new safety concerns. Tool-augmented judges may access external systems such as search engines, code executors, or databases, which expands the attack surface for prompt injection, tool misuse, or unintended side effects. Multi-agent collaboration can further amplify risks if unsafe behaviors propagate across agents or if adversarial interactions emerge. Moreover, when judge agents are used to provide reward signals for model optimization, systematic biases or errors in agentic judgment may be reinforced and amplified during training, leading to unintended model behaviors.

\paragraph{Privacy.}
\textit{Agent-as-a-Judge} also introduces privacy challenges, particularly in settings that involve persistent memory or personalized evaluation. To maintain consistency or adapt judgments to specific users or contexts, judge agents may store intermediate states, user information, or historical interaction data. If not carefully designed, such memory mechanisms can increase the risk of sensitive data leakage or unauthorized inference about user attributes. This issue becomes more pronounced in professional domains such as medicine, law, or education, where evaluation often relies on confidential or personally identifiable information.

\subsection{Future Directions}

\paragraph{Personalization.}
Current \textit{Agent-as-a-Judge} systems are constrained by static, one-size-fits-all evaluation criteria, failing to align with diverse individual preferences. To bridge this gap, future research should focus on enhancing the autonomy and adaptivity of judge agents. A critical enabler is proactive memory management: rather than passively retrieving history, agents must actively manage the {lifecycle} of user-specific knowledge—autonomously deciding when to register new preferences, update evolving standards, or prune obsolete feedback. This agentic control transforms memory into a dynamic belief system, allowing the judge to continuously refine its criteria and maintain alignment with the user's specific values and usage contexts.

\paragraph{Generalization.}
Current systems rely on predefined rubrics constructed offline, limiting their ability to generalize across diverse or open-ended tasks. Future judge agents should leverage planning capabilities to dynamically discover and adapt evaluation criteria.
1) {Context-Aware Rubric Generation}: Agents should synthesize evaluation criteria on-the-fly by analyzing the specific intent and complexity of responses, identifying relevant assessment dimensions not anticipated during design.
2) {Adaptive Multi-Granularity Scoring}: Rubrics should dynamically scale based on task difficulty—applying high-level holistic criteria for straightforward tasks, while decomposing into fine-grained sub-rubrics for complex workflows.

\paragraph{Interactivity.}
Current systems operate as passive, one-way observers. Future agents should evolve into interactive evaluators that actively engage with both the environment and human stakeholders.
1) {Interactive Environmental Feedback}:
Instead of static test suites, judge agents should dynamically tailor evaluation trajectories—autonomously escalating task complexity or isolating edge cases to rigorously probe the evaluand's failure boundaries.
2) {Human-Agent Collaborative Calibration}:
To address subjective or ambiguity-rich scenarios, agents should leverage human-in-the-loop mechanisms. By proactively consulting experts to verify intent or resolve conflicts, the judge refines its criteria through multi-turn alignment, ensuring higher trust and interpretability.

\paragraph{Optimization.} 
Current approaches predominantly rely on inference-time engineering, which is fundamentally bottlenecked by the fixed capabilities of frozen backbones. To transcend these limits, the field must pivot towards Training-based Optimization. This paradigm shift entails two key levels: 1) {Individual Capability}: Utilizing Reinforcement Learning (RL) to internalize complex agentic behaviors—such as sequential planning and adaptive tool use—that are difficult to elicit via prompting alone. 2) {Learned Coordination}: Extending optimization to multi-agent settings. Rather than ad-hoc inference collaboration, agents should be trained with joint objectives to intrinsically learn effective communication and consensus strategies.

\paragraph{Concluding Remarks: Towards True Autonomy.} As characterized in Section~\ref{sec:preliminary}, existing implementations exhibit varying degrees of agency.
The future directions discussed above—personalization, generalization, interactivity, and optimization—collectively point towards an evolutionary trajectory towards autonomy. The next generation of judge agents must transcend fixed protocols to become genuinely \textit{agentic} entities capable of self-directed adaptation, active context curation, and continuous self-refinement, ultimately realizing the full potential of agents that actively perceive, reason, and evolve alongside the models they assess.

%% file: 6_conclusion.tex
\section{Conclusion}

This paper provides the first comprehensive survey of \textit{Agent-as-a-Judge}. We established a novel taxonomy and demonstrated how agentic capabilities, including multi-agent collaboration, autonomous planning, tool integration, and memory, overcome the limitations of naive LLM judges to deliver more robust, verifiable and nuanced judgments across general and professional domains.
While promising, this evolution presents challenges in computational cost, latency, safety, and privacy. Future progress should prioritize personalization, generalization, and optimization, ultimately realizing truly autonomous evaluators that continuously adapt to the evolving AI landscape.

\section*{Limitations}

\paragraph{Early Stage of Paradigm Consensus.} As a pioneering survey exploring the evolution of \textit{Agent-as-a-Judge}, this study faces the challenge that the field has not yet gained complete widespread recognition in academia. Although the transition from \textit{LLM-as-a-Judge} to \textit{Agent-as-a-Judge} has begun to take shape, there is still a lack of long-term consensus regarding the definition of evaluation agents. 
Nevertheless, establishing this foundational framework is essential to orienting future research. We are committed to iteratively refining this taxonomy as the paradigm matures and gains broader recognition.

\paragraph{Inclusion of Early Prompting Methods.}
We acknowledge a potential gap between early methodologies and the increasingly rigorous definitions of agents. Many pioneering works in automated evaluation, though named as "agent",  rely heavily on prompting engineering, such as fixed role-play, which may not align with the strict criteria for autonomy, dynamic planning, or tool-use held by the current community.
Nevertheless, we deliberately include these prompt-based frameworks as they represent the initial shift from monolithic inference toward dynamic decomposition and self-evolving systems. Excluding them would obscure the transition thus compromising a complete understanding of the field's evolution.


\section*{Ethics Statement}
This work does not involve the use or creation of datasets or scientific artifacts that would require specific ethical clearance, data privacy considerations, or licensing agreements. We believe this work adheres to the ethical guidelines of the conference and poses no immediate negative social impact.